\pdfoutput=1

\documentclass[11pt]{article}

\usepackage[]{EMNLP2023}

\usepackage{hyperref}
\usepackage{pifont}

\usepackage{amssymb}
\usepackage{mathtools}
\usepackage{amsthm}

\usepackage[capitalize,noabbrev]{cleveref}

\usepackage[textsize=tiny]{todonotes}

\usepackage{ulem}

\makeatletter
\def\adl@drawiv#1#2#3{%
        \hskip.5\tabcolsep
        \xleaders#3{#2.5\@tempdimb #1{1}#2.5\@tempdimb}%
                #2\z@ plus1fil minus1fil\relax
        \hskip.5\tabcolsep}
\newcommand{\cdashlinelr}[1]{%
  \noalign{\vskip\aboverulesep
           \global\let\@dashdrawstore\adl@draw
           \global\let\adl@draw\adl@drawiv}
  \cdashline{#1}
  \noalign{\global\let\adl@draw\@dashdrawstore
           \vskip\belowrulesep}}
\makeatother

\newcommand{\cmmnt}[1][]{\ignorespaces}
\newcommand{\eqcolor}[1]{\textcolor{blue}{#1}}
\newcommand{\makeinvisible}[1]{}

\newcommand{\glue}[0]{GLUE}

\newcommand{\token}[0]{TOKEN}

\newcommand{\standard}[0]{PLMs}
\newcommand{\datamux}[0]{DataMUX}  
\newcommand{\tmux}[0]{T-MUX}  
\newcommand{\mux}[0]{MUX}
\newcommand{\demux}[0]{DeMUX}

\newcommand{\muxedrepresentation}[0]{$\mathbf{x}^{\textrm{\texttt{{\small \mux{}}}}}$}
\newcommand{\muxedoutput}[0]{$\mathbf{h}^{\textrm{\texttt{{\small \mux{}}}}}$}

\newcommand{\electra}[0]{ELECTRA}
\newcommand{\bert}[0]{BERT}
\newcommand{\muxelectra}[0]{MUX-ELECTRA}
\newcommand{\muxbert}[0]{MUX-BERT}
\newcommand{\muxelectratab}[0]{MUX-ELEC}

\newcommand{\newdemux}[0]{RSA-\demux{}}
\newcommand{\newmux}[0]{Contextual}
\newcommand{\mainmux}[0]{Non-contextual}
\newcommand{\muxplm}[0]{MUX-PLM}
\newcommand{\muxplms}[0]{MUX-PLMs}
\usepackage{amsfonts,amsmath}
\usepackage{times}
\usepackage{latexsym}
\usepackage[T1]{fontenc}
\usepackage[utf8]{inputenc}
\usepackage{microtype}
\usepackage{booktabs,arydshln}
\usepackage{multirow}
\usepackage{graphicx}
\usepackage{caption}
\usepackage{subcaption}

\title{MUX-PLMs: Data Multiplexing for High-throughput Language Models}

\author{
  \textbf{Vishvak Murahari$^{1}$ \qquad Ameet Deshpande$^{1}$ \qquad Carlos E. Jimenez$^{1}$  } \\
  \textbf{Izhak Shafran$^{2}$ \qquad Mingqiu Wang$^{2}$ \qquad Yuan Cao$^{2}$ \qquad Karthik Narasimhan$^{1}$} \\\\
  $^{1}$Princeton University \qquad $^{2}$ Google Brain \\
  \texttt{murahari@cs.princeton.edu}
}

\begin{document}
\maketitle

\begin{abstract}

The widespread adoption of large language models such as ChatGPT and Bard has led to unprecedented demand for these technologies.
The burgeoning cost of inference for ever-increasing model sizes coupled with hardware shortages has limited affordable access and poses a pressing need for efficiency approaches geared towards high throughput and performance.
Multi-input multi-output (MIMO) algorithms such as data multiplexing, offer a promising solution with a many-fold increase in throughput by performing inference for multiple inputs at the cost of a single input.
Yet these approaches are not currently performant enough to be deployed in modern systems. We change that by developing MUX-PLMs, a class of high throughput pre-trained language models (PLMs) trained with data multiplexing, that can be fine-tuned for any downstream task to yield high-throughput high-performance. Our novel multiplexing and demultiplexing modules proficiently entangle and disentangle inputs, and enable high-performance high throughput \muxplms{} that are competitive with vanilla PLMs while achieving 2x/5x inference speedup with only a $1-4 \%$ drop on a broad suite of tasks.
\footnote{Code + Models: \url{https://github.com/princeton-nlp/datamux-pretraining/}.}

\end{abstract}

\section{Introduction}
\label{sec:introduction} 

\begin{figure*}[t!]
    \centering
    \includegraphics[width=\textwidth]{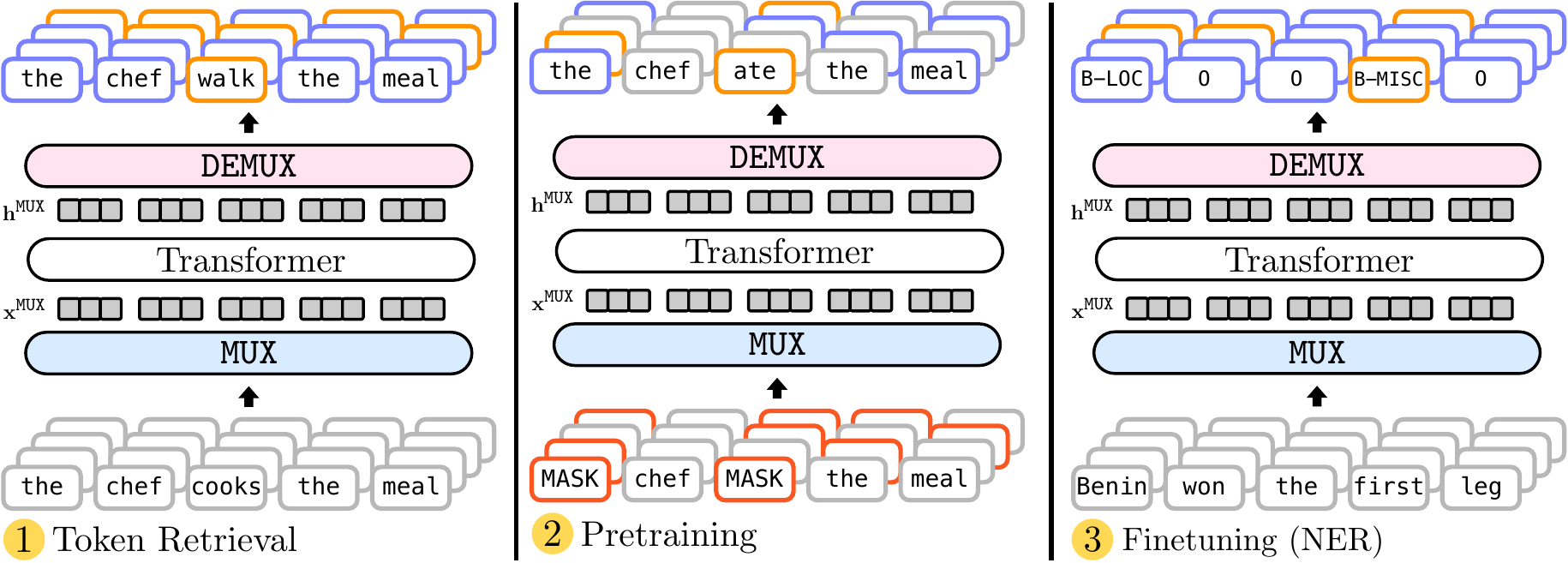}
    \caption{
    Illustrating the training process for \muxplms{}.
    \muxplms{} are first primed for MIMO style training with a token-retrieval auto-encoding task, where the model is trained to output the tokens in the $N$ inputs.
    \muxplms{} are then pre-trained by adapting standard pre-training objectives (\bert{} in this example), to MIMO style training with data multiplexing. 
    The resulting \muxbert{} model, similar to standard PLMs, provides a general model initialization that can be fine-tuned on any downstream task (NER in this example). 
    Output predictions are shown above the system head with highlighted predictions contributing to the gradient update; violet indicates a correct prediction while orange indicates an incorrect prediction. 
    Red highlighted tokens in the input indicate a position that has been masked.}
    \label{fig:training_stages}
\end{figure*}

Language models like ChatGPT~\cite{chatgpt}, PaLM~\cite{chowdhery2022palm}, T5~\cite{raffel2020exploring}, and CM3~\cite{aghajanyan2022cm3}, have seen unprecedented adoption in diverse sectors ranging from education and healthcare to manufacturing and marketing. 
The proficiency of these tools has led to unprecedented demand for these models, with users facing frequent outages and capacity limits.
Additionally, ever-increasing model sizes and hardware shortages have constrained models' ability to handle a very high load of requests, thus limiting large-scale affordable access to these models.
These trends bring into focus the need for high-throughput, high-performance, efficient, and environmentally responsible models that can be deployed at scale to meet the quickly growing demand.

Multi-input Multi-output architectures (MIMO) \cite{havasi2021training, Rame_2021_ICCV,datamuxmurahari} are a promising hardware-agnostic and architecture-agnostic paradigm that perform inference for multiple inputs \textit{simultaneously} at the cost of a single input. 
This efficiency paradigm is natively geared towards yielding high-throughput models, in addition to being complementary in approach and motivation to current efficiency methods such as pruning, quantization, and distillation.
Interestingly, MIMO approaches are partly inspired by the human brain's extraordinary ability to process multiple inputs and propagate information at a high bandwidth with a few neural codes~\cite{blumhagen2011neuronal,akam2014oscillatory,pirschel2016multiplexed,hong2016multiplexed,friedrich2004multiplexing}. 

\citet{datamuxmurahari} introduced data multiplexing, a MIMO technique that can enable a many-fold increase in throughput. The method compresses $N$ different instances into a single ``multiplexed'' hidden representation before decompressing it into $N$ independent predictions.
While they show the plausibility of MIMO training, their method leads to a significant drop in performance ($20-30\%$ points) compared to state-of-the-art models.


In this work, we introduce MUX-PLMs, a class of high-throughput pre-trained language models trained in a MIMO fashion with data multiplexing to process multiple inputs (2-10) simultaneously with a forward pass over a single instance. 
\muxplms{} offer up to $400\%$ improvement in throughput over baseline pre-trained models while only being $\sim4$ points and $\sim2$ points worse than baseline pre-trained language models for text classification and token classification tasks, respectively. 
\muxplms{}, like other pre-trained language models, provide general model initialization that can be fine-tuned for \textit{any} downstream task.
We demonstrate the effectiveness and generality of our \muxplms{} class of pre-trained models by training \muxbert{} and \muxelectra{} models, which are trained with pre-trained objectives adapted from \bert{}~\cite{devlin-etal-2019-bert} and \electra{}~\cite{clark2020electra} respectively, although in a MIMO fashion with data multiplexing.

Our work is the first to introduce MIMO architectures to PLMs. To enable this, we first develop a new demultiplexing module, RSA-demux (Figure~\ref{fig:electra_schematic}), that randomly initializes and learns private key vectors to recover the multiple outputs from a multiplexed representation. 
Secondly, we introduce a new \textit{Contextual Multiplexer} module (Figure~\ref{fig:attention_multiplexing}) that uses a cross-instance attention-based mechanism to aggregate context across the set of multiplexed instances, which seems to be particularly effective for token-level tasks. 
Thirdly, our three-stage training algorithm (Figure~\ref{fig:training_stages}) enables stable and efficient training of MUX-PLMs.

Importantly, \muxplms{} are complementary to existing state-of-the-art model compression techniques.
We hope our work validates MIMO architectures as a promising complementary direction to existing efficiency techniques. 
Consequently, we hope future research develops MIMO architectures in tandem with other efficiency approaches, leveraging the best of both paradigms. 
We publicly release our models and code to promote open-source research on the next generation of MIMO architectures for large language models.

\section{Related Work}
\label{sec:related}

\paragraph{Efficient Inference with Transformers}
Recent methods in NLP rely heavily on transfer learning through Transformer-based~\cite{Vaswani2017AttentionIA} language models trained on large text corpora using self-supervised objectives, such as autoregressive~\cite{Radford2018ImprovingLU} or masked language modeling~\cite{devlin-etal-2019-bert}. 
Prior analysis on pre-training language models has observed power-law scaling of model performance with respect to model size~\cite{kaplan_scaling_laws}, leading the community to develop ever-larger language models. 
 It is also generally recognized that pre-trained language models are significantly over-parameterized; effectively learning a \textit{subnetwork} that utilizes only a relatively small number of their total parameters~\cite{voita-etal-2019-analyzing, NEURIPS2019_2c601ad9, gordon-etal-2020-compressing, prasanna-etal-2020-bert}.

The ubiquity of pre-trained language models, their growing size, and over-parameterization has inspired extensive research on improving inference efficiency. 
This includes methods such as structured pruning~\cite{liu2018rethinking, Wang_2020, lagunas2021block, cofi, yang-etal-2022-textpruner}, knowledge distillation~\cite{Hinton2015DistillingTK, sanh2019distilbert, sun2020mobilebert, jiao2020tinybert, yin2021autotinybert}, quantization~\cite{Zafrir2019Q8BERTQ8, Shen2020QBERTHB}, and data multiplexing~\cite{datamuxmurahari}.
These approaches assume that PLMs are highly over-parametrized and attempt to approximate a large function by learning a smaller, compressed, version of the original model. 
Past work has also focused on unstructured pruning for both task finetuning~\cite{NEURIPS2020_b6af2c97, NEURIPS2020_eae15aab} and pre-trained~\cite{zafrir2021prune, https://doi.org/10.48550/arxiv.2210.06210} language model settings, but don't increase model throughput due to hardware limits.

\paragraph{Multi-input Multi-output Models}
While pruning, quantization, and distillation seek to reduce overparameterization by reducing models' representational capacity, other lines of work seek to exploit overparameterization in other ways. 
Multi-input Multi-output (MIMO) architectures~\cite{havasi2021training, Rame_2021_ICCV, datamuxmurahari} train models using mixed-instance representations, i.e. ~\citet{zhang2018mixup}, in order to obtain predictions for multiple instances simultaneously. 
Unlike efficiency methods, ~\citet{havasi2021training} and ~\citet{Rame_2021_ICCV} try to obtain better performance by inducing multiple subnetworks in a single convolutional model to perform ``ensembling for free'' during inference. 
Data multiplexing, introduced in \datamux{} \cite{datamuxmurahari}, aims to improve model efficiency by training Transformer models with mixed-instance representations to perform simultaneous inference for language tasks, thereby improving inference throughput many-fold. 
Currently, MIMO architectures have only been used in a limited setting, achieving middling performance. 
Our work training PLMs with data multiplexing, dramatically improves inference throughput while preserving high accuracy for downstream tasks.

\section{Methodology}
\label{sec:methodology}

\begin{figure*}[!t]
    \centering
\includegraphics[width=0.85\textwidth]{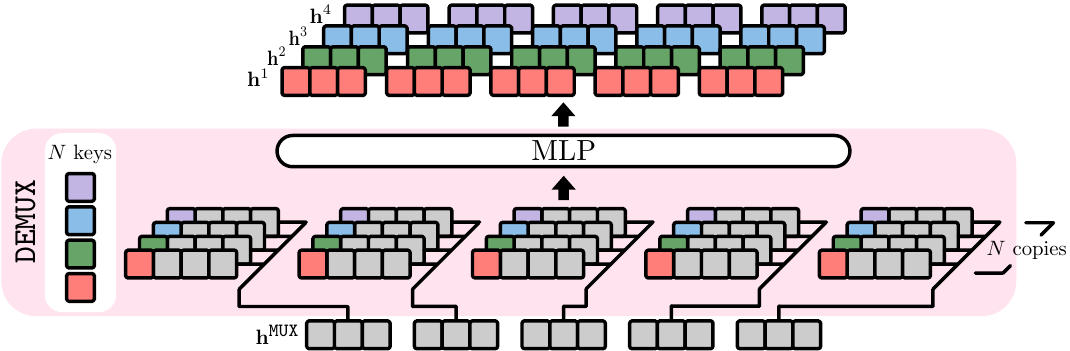}
    \caption{Illustrating our novel RSA-inspired demultiplexing module. The module is initialized with N key vectors which are used to demultiplex the transformed multiplexed representations ($h^{MUX}$). The keys are concatenated with $h^{MUX}$ and are processed with an MLP to generate the demultiplexed output representations ($h_1 \cdots h_4$).}
    \label{fig:electra_schematic}
\end{figure*}

\begin{figure}[ht]
    \centering
    \includegraphics{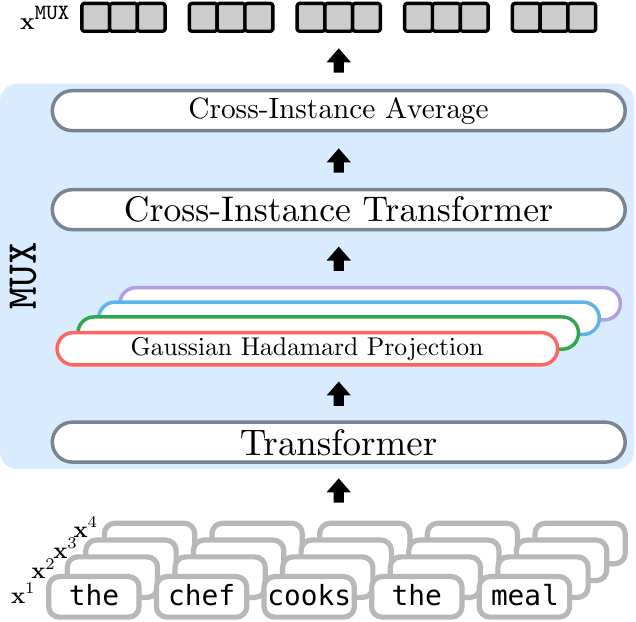}
    \caption{Illustrating our attention-based multiplexing module. The module generates contextual representations for instances $x_1 \cdots x_4$ with a Transformer layer and then applies a hadamard product between the contextual representations and the corresponding multivariate gaussian to generate instance-conditioned representations. The final multiplexed representations are generated by first applying another Transformer layer, which attends across the instances for all the positions in the sequence, and then averaging across the instances.}
    \label{fig:attention_multiplexing}
\end{figure}

We briefly introduce readers to the data multiplexing MIMO architecture~\cite{datamuxmurahari}, which we denote \tmux{}.
We then detail our novel approach to train \muxplms{} to yield high-throughput and performant language models.

\subsection{\tmux{}: Data multiplexing with Transformer}
\label{sec:methodology:datamux}

Data multiplexing allows parallel processing of multiple sequences with a single forward or backward pass through the model ($M$) and requires two crucial components, multiplexer, and demultiplexer.

\paragraph{Multiplexer}
The multiplexer module (\mux{}) combines an ordered set of multiple inputs -- $X^{1:N} = \left ( \mathbf{x}^1, \dots, \mathbf{x}^N \right )$ into a single superimposed representation (\muxedrepresentation{}).
If $\mathbf{x}^{i} \in \mathbb{R}^d$, the multiplexer is a transformation ($\textrm{MUX} \colon \mathbb{R}^{N\times d} \to \mathbb{R}^d$) such that $\textrm{\muxedrepresentation{}} = \textrm{\mux{}}\left ( X^{1:N} \right )$.

To ensure \mux{} is an order-preserving transformation,\tmux{} samples a vector ($\mathbf{v}^i \in \mathbb{R}^d$) from a standard multivariate Gaussian and applies the Hadamard product (element-wise multiplication) with the corresponding input representation ($\mathbf{x}^i$) before summing up vectors for all positions.
\begin{align}
    \begin{split}
        \textrm{\muxedrepresentation{}} &= \textrm{\mux{}}\left ( X^{1:N} \right ) = \dfrac{1}{N} \sum_{i=1}^N \mathbf{x}^i \odot \mathbf{v}^i \\
        \mathbf{v}^i &\in \mathbb{R}^d \sim \mathcal{N}\left ( \mathbf{0}, \mathbf{I} \right )   \\
    \end{split}
\end{align}
The model processes the multiplexed representation and emits a multiplexed hidden state -- $\textrm{\muxedoutput{}} = M\left ( \textrm{\muxedrepresentation{}} \right )$.
To multiplex Transformers' sequenced inputs $\left (\mathbf{x}^i = \left ( \mathbf{x}^i_{1}, \dots, \mathbf{x}^i_{L}  \right ) \right )$ of length $L$,
\tmux{} applies the same $\mathbf{v}^i$ to all $L$ positions of instance $i$.
\begin{align}
    \begin{split}
        \textrm{\muxedrepresentation{}} &= \textrm{\mux{}}\left ( X^{1:N} \right ) =\\
        & \left ( \dfrac{1}{N} \sum_{i=1}^N \mathbf{x}^i_1 \odot \mathbf{v}^i, \dots, \dfrac{1}{N} \sum_{i=1}^N \mathbf{x}^i_L \odot \mathbf{v}^i \right )
    \end{split}
\end{align}

\paragraph{Demultiplexer}
A prediction needs to be made for each instance in $X^{1:N}$, and~\tmux{}'s demultiplexer module (\demux{}) achieves this by separating the superimposed output (\muxedoutput{}) into $N$ output representations corresponding to the input ($\mathbf{h}^1, \dots, \mathbf{h}^N$).

\begin{align}
    \begin{split}
        \mathbf{h}^i &= \textrm{\demux{}}\left ( \textrm{\muxedoutput{}}, \mathbf{p}^i \right ) \\
        \mathbf{h}^i_j &= \textrm{\demux{}}\left ( \textrm{\muxedoutput{}}_j, \mathbf{p}^i \right ) \\
    \end{split}
\end{align}

The vector $\mathbf{p}^i \in \mathbb{R}^d$ is dynamically generated for each instance ($i$) with the help of a prefix that is added to the input and re-used for all positions in the instance.
They add a $\textrm{\textit{prefix}}_i$ to $\mathbf{x}^i$, represented by the following pattern, where $\epsilon^i$ is a special token, and $\mathbf{p}^i$ is set to be the output corresponding to token $i$ in the prefix.
\begin{align*}
    &\textit{prefix}^1 \hspace{0.3em}= [\eqcolor{\epsilon^1}, \epsilon^{\text{pad}}, \ldots, \epsilon^{\text{pad}} ] \\
    &\textit{prefix}^2 \hspace{0.3em}= [\epsilon^{\text{pad}}, \eqcolor{\epsilon^2}, \epsilon^{\text{pad}}, \ldots, \epsilon^{\text{pad}} ] \\
    &\cdots \\
    &\textit{prefix}^N = [\epsilon^{\text{pad}}, \ldots, \epsilon^{\text{pad}}, \eqcolor{{\epsilon^N}} ]
\end{align*}

\subsection{\muxplms{}: Data multiplexing for high-throughput language models}
We propose \muxplms{}, a class of high-throughput pre-trained Transformer-based language models trained in a MIMO fashion with data multiplexing.  
To demonstrate the viability and the generality of this class of models, we pre-train Transformer models with objectives based on \bert{} and \electra{}, to get \muxbert{} and \muxelectra{} respectively.
\muxplms{} are trained with our three stage training algorithm (Figure~\ref{fig:training_stages}). 
Firstly, \muxplms{} are trained with the token retrieval task in \tmux{}, which is an auto-encoding setup to decode all the tokens in the multiplexed input. 
This simple auto-encoding task is critical to prime the model for MIMO-style data multiplexing.
The MUX-PLMs are then pre-trained with standard pre-training objectives but adapted to MIMO-fashioned training with data multiplexing. 
MUX-PLMs show significant throughput improvement over standard pre-trained LMs while matching their downstream task accuracies. 
Finally, \muxplms{}, like other pre-trained language models, provide general model initialization that can be fine-tuned for \textit{any} downstream task.

\paragraph{Contextual multiplexer}

\tmux{}'s multiplexer multiplexes tokens independent of 1) tokens in the same position in other instances and 2) other tokens in the instance, which could lead to suboptimal representations.
We, therefore, explore a contextual multiplexing scheme that aggregates context both from tokens in the same instance and tokens in the same position of other instances (Figure~\ref{fig:attention_multiplexing}).
We first use a single transformer layer $\textsc{TRANS}_{\text{ctx}}$ to get contextual representations
$\mathbf{h}_{\text{ctx}}^i = \textsc{TRANS}_{\text{ctx}} \left ( \mathbf{x}^i_{1}, \dots, \mathbf{x}^i_{L}  \right )$) of length $L$.
We apply a hadamard product with a multivariate gaussian $\mathbf{v}^i$ to all $L$ positions.
\begin{align}
    \mathbf{g}_{\text{ctx}}^i= \mathbf{h}_{\text{ctx}}^i \odot \mathbf{v}^i
\end{align}
We generate multiplexed representations, \textrm{\muxedrepresentation{}}, by applying another transformer layer $\textsc{TRANS}_{\text{inst}}$ across encoded representations from $N$ instances at each position from $1$ to $L$. This is done by transposing  $\mathbf{g}_{\text{ctx}}$ and applying $\textsc{TRANS}_{\text{inst}}$.
\begin{align}
    \textrm{\muxedrepresentation{}}= \textsc{TRANS}_{\text{inst}} \left (\mathbf{g}_{\text{ctx}}^{\top} \right)
\end{align}

\paragraph{RSA demultiplexer}

The \textit{demultiplexer} in \tmux{} requires a prefix whose length scales linearly with the number of instances ($N$), thus reducing the effective context length during pre-training, which degrades performance~\cite{DBLP:journals/corr/abs-2004-08483,zaheer2020big,beltagy2020longformer}.
Furthermore, it decreases throughput during inference for large $N$ because the model must process an extra prefix of length $N$ for each of the $N$ instances.
To address these issues, we draw inspiration from the RSA cryptosystem~\cite{rivest1978method} to randomly initialize and learn $N$ (private) key vectors $(\mathbf{k}_1, \dots, \mathbf{k}_N$, $\mathbf{k}_i \in \mathbb{R}^d)$ which are keys that can be used to demultiplex the output representation (Figure~\ref{fig:electra_schematic}).

\begin{align}
    \begin{split}
        \mathbf{h}^i &= \textrm{\demux{}}\left ( \textrm{\muxedoutput{}}, \mathbf{k}^i \right ) \\
        \mathbf{h}^i_j &= \textrm{\demux{}}\left ( \textrm{\muxedoutput{}}_j, \mathbf{k}^i \right ) \\
    \end{split}
\end{align}

Akin to RSA, $\mathbf{v_i}$ and $\mathbf{k_i}$ can be treated as the keys for multiplexing (encryption) and demultiplexing (decryption) while ensuring that, unlike \tmux{}, the input sequence length does not change and thereby leading to an improvement in throughput. 
Importantly, this architecture ensures that the keys better transfer across the different stages of training as they are no longer conditioned on the input instances. 

\section{Experimental Setup}
\label{sec:experimental}

\begin{table}[t]
\centering
\resizebox{\columnwidth}{!}{%
\begin{tabular}{@{}llccccc@{}}
\toprule
\multirow{2}{*}{\textbf{Model}} &
\multirow{2}{*}{$\mathbf{N}$} &
\multicolumn{2}{c}{\textbf{GLUE}} &
\multicolumn{2}{c}{\textbf{Token}} &
\multirow{2}{*}{$\nearrow$}
\\ \cmidrule(lr){3-4} \cmidrule(lr){5-6}
    &
    &
    Mean (std) &
    Max &
    Mean (std)&
    Max & \\ \midrule
\textbf{\bert{}} & \multirow{2}{*}{\textbf{1}}  & \textbf{85.4 (0.0)} & \textbf{85.4} & \textbf{95.8 (0.0)} & \textbf{95.8} & 1.0$\times$ \\
\textbf{\electra{}} & & 82.1 (0.0) & 82.1 & 95.3 (0.0) & 95.3 & 1.0$\times$ \\
\midrule
\textbf{\tmux{}} & \multirow{3}{*}{\textbf{2}}  & 60.4 (0.6) & 61.8 & 81.4 (0.1) & 81.5 & 1.9$\times$  \\
\textbf{\muxbert{}}$^\ddagger$ & & \textbf{82.5 (0.6)} & \textbf{83.4} & \textbf{95.2 (0.1)} & \textbf{95.4} & \textbf{2.0}$\times$\\
\textbf{\muxelectratab{}$^\ddagger$} & & \textbf{82.5 (0.4)} & 83.1 & 95.0 (0.0) & 95.1 & \textbf{2.0}$\times$\\
\midrule
\textbf{\tmux{}} & \multirow{3}{*}{\textbf{5}} & 59.7 (0.6) & 60.6 & 81.3 (0.2) & 81.5 & 4.4$\times$   \\
\textbf{\muxbert{}$^\ddagger$} & & \textbf{80.3 (0.4)} & \textbf{80.9} & \textbf{93.6 (0.1)} & \textbf{93.6} & \textbf{4.9}$\times$\\
\textbf{\muxelectratab{}$^\ddagger$} & & 79.8 (0.6) & 80.5 & 93.4 (0.0) & 93.5 &  \textbf{4.9}$\times$\\
\midrule
\textbf{\tmux{}} & \multirow{3}{*}{\textbf{10}}  & 58.1 (0.5)  & 59.1 & 79.7 (0.2) & 80.0 & 8.4$\times$  \\
\textbf{\muxbert{}}$^\ddagger$ & & 77.8 (0.6) & 78.8 & 91.6 (0.1) & \textbf{91.8} & \textbf{9.8}$\times$\\
\textbf{\muxelectratab{}}$^\ddagger$ & & \textbf{78.2 (0.6)} & \textbf{79.0} & \textbf{91.7 (0.1)} & \textbf{91.8} & 9.7$\times$\\
\bottomrule
\end{tabular}
}
\caption{Average GLUE and token-level classification scores for the \textsc{base} (L=12, H=768) configuration, across ELECTRA, BERT, and MUX-PLMs for $N\in\{1,2,5,10\}$. $\ddagger$ indicates our models and $\nearrow$ indicates throughput increase w.r.t. to a vanilla BERT$_{\textsc{base}}$ model. All models are evaluated on 5 seeds with mean and max scores reported.  
}
\label{tab:main_results}
\vspace{-10pt}
\end{table}
\paragraph{Datasets}
We pre-train all models on Wikipedia~\cite{wikidump} and Bookscorpus~\cite{bookscorpus}.
We evaluate on all datasets from the GLUE benchmark~\cite{wang2018glue}, which are CoLA~\cite{colawarstadt2019neural}, SST-2~\cite{sstsocher2013recursive}, MRPC~\cite{mrpcdolan2005automatically}, QQP~\cite{qqp}, STS-B~\cite{stsbcer2017semeval}, MNLI~\cite{mnliwilliams2018broad}, QNLI~\cite{wang2018glue}, RTE~\cite{wang2018glue}, and WNLI~\cite{wnlilevesque2012winograd}. 
We also evaluate on token classification tasks -- named entity recognition~\cite{Sang2003IntroductionTT} and POS tagging~\cite{grunewald2021coordinate}. 
We don't report average over the two smallest tasks in GLUE, WNLI and CoLA, as we observe high variance across seeds. 
All numbers are reported on the dev split. We report scores for all tasks in Appendix~\ref{appendix:taskperf}.

\paragraph{Models} We experiment with ELECTRA and BERT pre-training objectives and present the pre-trained multiplexed models \textbf{\muxbert{}} and \textbf{\muxelectra{}} for $N =\ 2,5 \text{ and } 10$. To simplify training, we use a random generator to train \muxelectra{} models, presented as an ablation in~\citet{clark2020electra}, instead of using a smaller masked LM. Except where otherwise noted, we do not use the contextual \mux{} module, but instead, use the RSA demultiplexing module. 
Refer to Appendix~\ref{appendix:pre}~and~\ref{appendix:fine} for implementation details.

\paragraph{Baselines} We run experiments to compare our models against \tmux{}, from~\citet{datamuxmurahari} and baseline PLMs - \electra{} and \bert{}, across three different model configurations (\textsc{small}, \textsc{base}, and \textsc{large}). 
We also provide a comparison to results reported in recent PLM pruning and distillation papers in Table~\ref{tab:pruning_distillation_baselines}.
\paragraph{Multi-run evaluation} We evaluate all models across $5$ random seeds to reduce variance for smaller datasets which is caused by the randomized order in which we multiplex instances in the batch.
We report both the average and maximum scores across seeds in Table~\ref{tab:main_results} to understand the importance of ordering the multiplexed instances and report average across seeds for all other results.

\section{Results}
\label{sec:results}

\subsection{Comparing \muxplms{} with \standard{} and \tmux{}}
Table~\ref{tab:main_results} shows that both \textbf{\muxbert{} and \muxelectra{} outperform \tmux{} at all levels of multiplexing ($N$)}, with improvements between $12$ and $20$ points on \glue{} and token-classification tasks respectively.
Furthermore, \muxplms{}' efficient RSA-inspired demultiplexing method allows it to achieve faster throughput than \tmux{}, increasing it by over $16\%$ for $N=10$.

Moreover, \textbf{\muxplms{} provide a significant boost in throughput ($N$ times faster) when compared to \standard{}, without a significant loss in performance.}
For example, \muxelectra{} ($N=2$) is $0.4$ points better and only $0.3$ points worse than \electra{} for \glue{} and \token{} tasks respectively, while being $2\times$ faster.
Similarly, \muxbert{} ($N=2$) is within $3$ and $0.6$ points of \bert{} for \glue{} and \token{} tasks respectively, while being significantly faster.
We also observe that as $N$ increases, \muxplms{}' throughput is significantly better, though performance compared to \standard{} can degrade.
This is because a large $N$ implies that \muxplms{} must parallelly process more instances, thus having to share network parameters and activations with a larger number of instances, thus improving throughput and degrading performance.
For example, the gap between \electra{} and \muxelectra{} on \token{} for $N=2$ is $0.2$ points and increases to $3.5$ points for $N=10$, which shows that $N$ serves as a parameter to control the performance-throughput trade-off. We explore this further in Section \ref{subsection:effect_of_varying_model_size} and Figure~\ref{fig:pareto_plots}.

\subsection{Comparing \muxplms{} with recent model compression methods}
\label{sec:results:baseline_comparison}

\begin{table}[!t]
    \centering
    \resizebox{0.9\columnwidth}{!}{%
    \begin{tabular}{llrrr}
    \toprule
   \textbf{Model} &  $\nearrow$ &  \textbf{QNLI} & \textbf{QQP} & \textbf{SST2}\\ 
    \midrule
    BERT &  $1.0\times$ &  90.5 & 91.2 & 91.7 \\
    \muxbert{} (N=2) & $2.0\times$ & 88.2 & 90.4 & 90.6 \\
    \muxbert{} (N=5) & $4.9\times$ & 85.6 & 88.8 & 86.9\\
    \midrule
    \multicolumn{5}{c}{\textbf{Use additional unlabelled or task-specific data}} \\
    DistilBERT$_6$ & ${2.0\times}$ & 89.2 & 88.5 & 91.3 \\
    Block Pruning & $2.7\times$ & 89.7 & - & 91.2 \\ 
    Prune OFA & $1.0\times$ & 90.3 & 91.2 & 91.5\\
    \hdashline
     \multicolumn{5}{l}{\textbf{Hybrid Approaches}} \\
    TinyBERT$_6$ &  ${2.0\times}$ & 91.1 & 91.1  & 93.0 \\
    CoFi & $2.7\times$ & 91.3 & - & 93.0 \\
    AutoTinyBERT  & $4.3\times$ & 89.7 & 89.9 & 91.4  \\
    MobileBERT & $2.3\times$ & 91.0 & - & 92.1 \\
    \bottomrule
    \end{tabular}
    }
    \caption{\muxplms{} are complementary to existing efficiency methods, while being competitive standalone.
    Contrary to existing methods, \muxplms{} \textit{do not use additional unlabelled and task-specific data} and can be easily fine-tuned for \textit{any} downstream task without architectural modifications.
    The inference speedups ($\nearrow$) are reported against BERT$_{\textsc{base}}$.
    }
    \label{tab:pruning_distillation_baselines}
    \vspace{-10pt}
\end{table}

We compare our \muxplm{} models with other efficient learning methods, such as pruning and distillation, in Table~\ref{tab:pruning_distillation_baselines}.
Contrary to other methods, our \textit{vanilla} \muxplms{} achieve competitive performance and significant throughput improvement without additional unlabeled and task-specific data, and can be easily fine-tuned to \textit{any} downstream task without any architectural modifications. 
For instance, when compared to DistilBERT, \muxbert{} ($N=2$) does $1$ point worse on QNLI and $2$ points better on QQP while being equally fast and not requiring any additional unlabeled data.

More broadly, methods like CoFi, AutoTinyBERT, and MobileBERT show that combining a wide range of paradigms (for example, CoFi combines structured pruning and knowledge distillation,  AutoTinyBERT combines knowledge distillation and neural architecture search, and MobileBERT combines knowledge distillation with novel architectural innovations) is a promising approach towards efficient high-performance models.

Towards this end, \muxplms{} are complementary in both approach and motivation to these methods, and can evolve in tandem with existing efficiency methods.
\muxplms{} demonstrate the viability of MIMO architectures for PLMs, in addition to being complementary to existing approaches, and we hope that MIMO architectures develop and evolve with other efficiency approaches while leveraging the best of all efficiency methods.

\subsection{Effect of varying model size}
\label{subsection:effect_of_varying_model_size}

\begin{table}[t]
\centering
\resizebox{0.8\columnwidth}{!}{
\begin{tabular}{@{}clccc@{}}
\toprule
\multicolumn{1}{l}{\textbf{Config}} & \textbf{Model} & \textbf{\glue{}} & \textbf{Token} & $\nearrow$ \\ \midrule
\multirow{3}{*}{\textsc{Small}} & \bert{}    &  80.6\makeinvisible{/80.6}                 &  94.0\makeinvisible{/94.0}             &  5.9$\times$  \\
& \tmux{}  & 59.5\makeinvisible{/60.2} & 81.8\makeinvisible{/82.0} & 8.7$\times$  \\
& \muxbert{}$^\ddagger$ &  79.0\makeinvisible{/79.8}                 &    93.3\makeinvisible{/93.4}             &  11.5$\times$   \\ \midrule
\multirow{3}{*}{\textsc{Base}}  & \bert{}  &   85.4\makeinvisible{/85.4}              &     95.8\makeinvisible{/95.8}           & 1.0$\times$   \\
& \tmux{}  &  60.4\makeinvisible{/61.8}                &         81.4\makeinvisible{/81.5}    &  1.9$\times$  \\
& \muxbert{}$^\ddagger$ & 82.5\makeinvisible{/83.4}                &     95.2\makeinvisible{/95.4}             & 2.0$\times$ \\ \midrule
\multirow{3}{*}{\textsc{Large}} & \bert{}        & 85.8\makeinvisible{/85.8}                 & 95.6\makeinvisible{/95.6}                 &   0.3$\times$ \\
& \tmux{}  & 61.7\makeinvisible{/62.2} & 80.9\makeinvisible{/81.1} & 0.6$\times$    \\
& \muxbert{}$^\ddagger$ &   84.1\makeinvisible{/84.6}     &    95.2\makeinvisible{/95.4}              & 0.6$\times$ \\
\bottomrule
\end{tabular}
}

\caption{
Changing the model size for \muxbert{} ($N=2$) models. Across different model sizes, \muxbert{} outperforms \tmux{} and achieve higher throughput (indicated under $\nearrow$ column). $\ddagger$ = our models.}
\label{tab:analysis_model_size}
\vspace{-10pt}
\end{table}
In this section, we show that \textbf{our multiplexing techniques work on a host of model sizes} and report results for \muxbert{} on three models sizes, \textsc{small}, \textsc{base}, and \textsc{large} for $N=2$ (Table~\ref{tab:analysis_model_size}).
We report results for other values of $N$ in the appendix.
\muxbert{}'s performance is close to that of \bert{} for all model sizes while having a significantly better throughput (the gap is less than $0.7$ points for \token{} tasks and $2.9$ points for \glue{} for close to twice the throughput). Multiplexing works effectively on all model sizes, with the drops with respect to \bert{} being $1.6$ and $1.7$ points on \glue{} for \textsc{Small} and \textsc{Large} respectively. \muxbert{}'s throughput is always $\approx 2\times$ that of \bert{}, which shows that a spectrum of \muxplm{} model sizes can be multiplexed during pre-training with competitive performance and with significantly higher throughput.

\begin{figure}[h]
  \centering
\begin{subfigure}{\columnwidth}
  \centering
  \includegraphics[width=0.85\columnwidth]{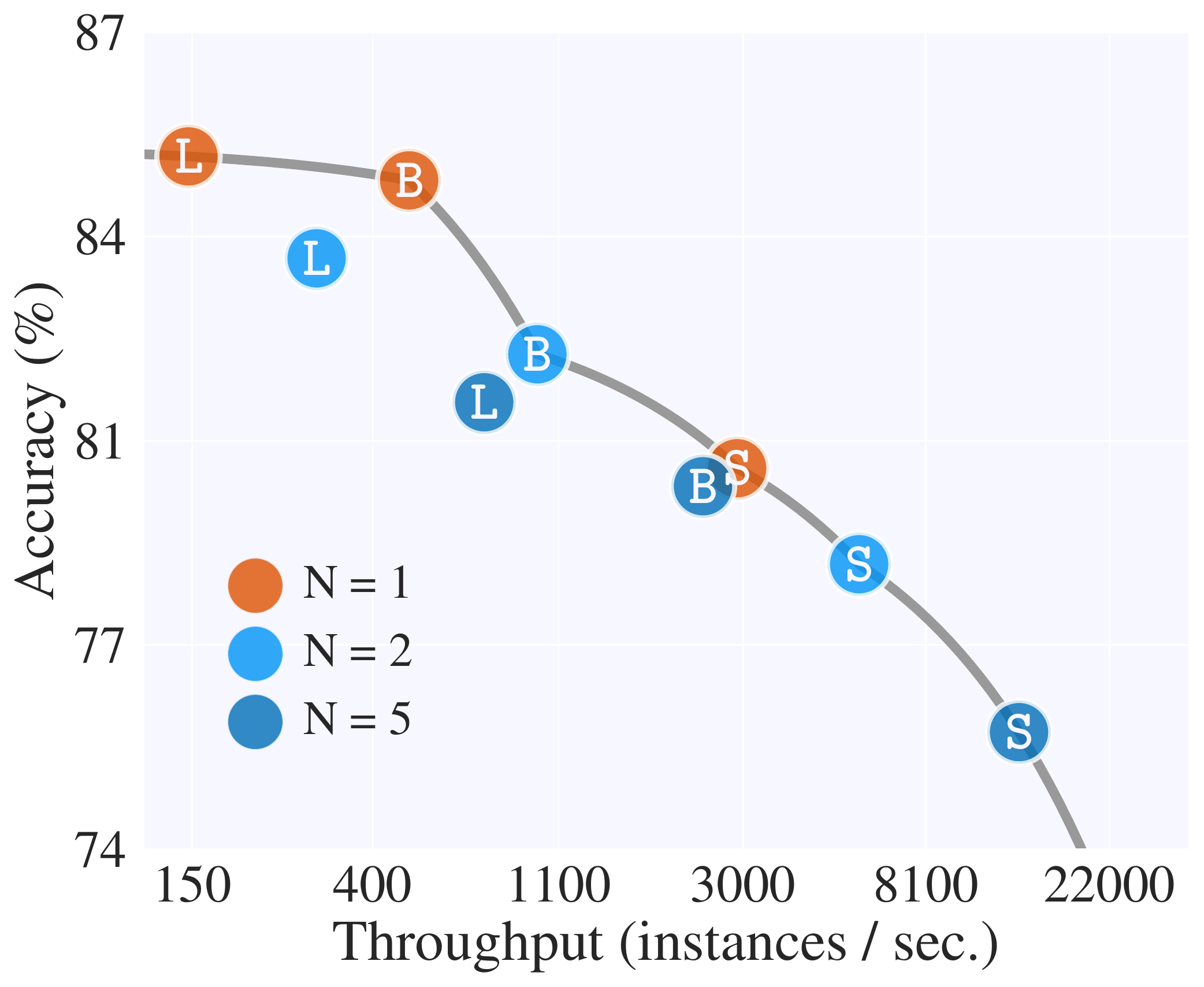}
\end{subfigure}\hfill
\begin{subfigure}{\columnwidth}
  \centering
  \includegraphics[width=0.85\columnwidth]{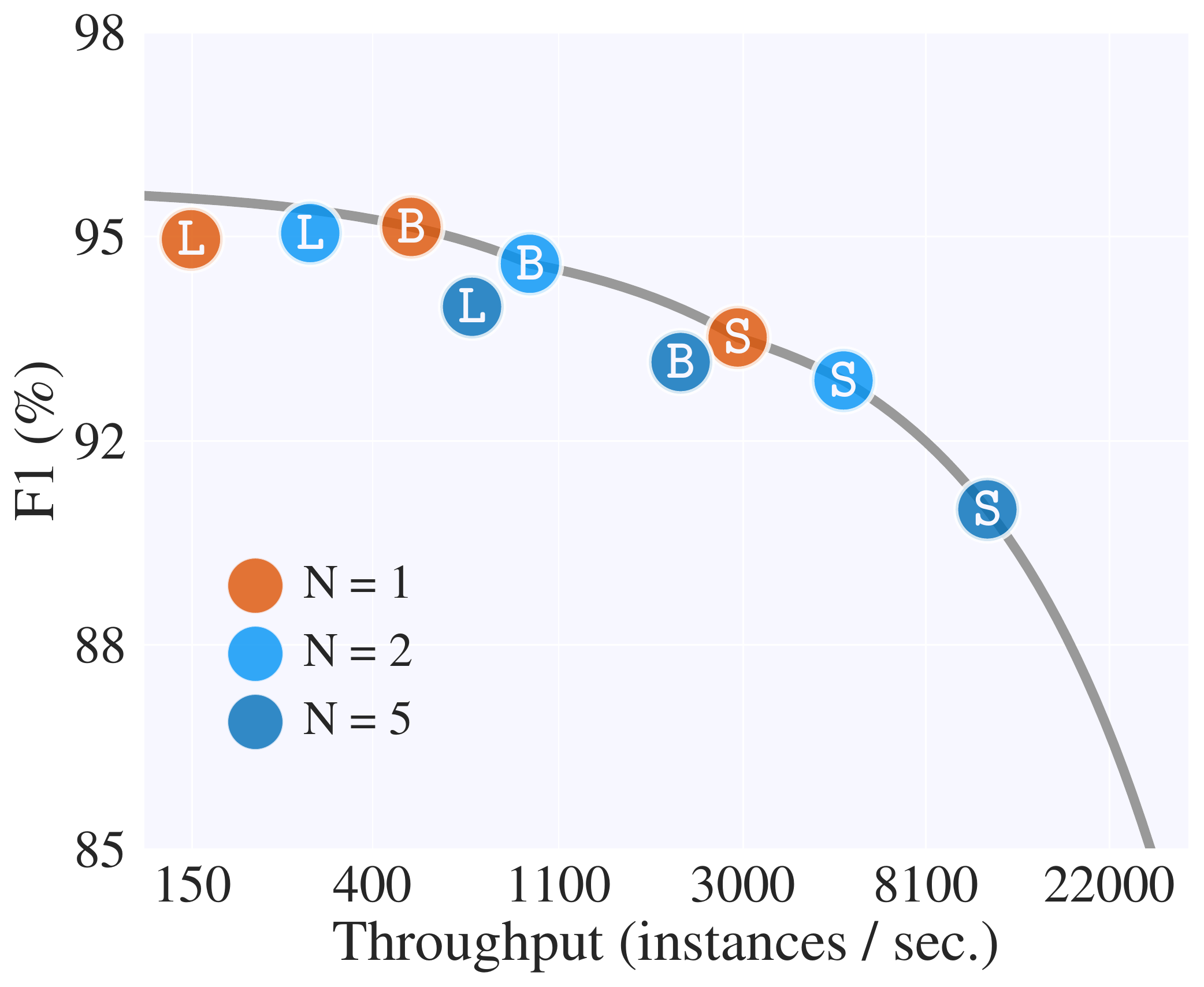}
\end{subfigure}
\caption{
(Top) BERT GLUE performance and throughput and 
(Bottom) BERT Token task performance and throughput, for $N\in\{1,2,5,10\}$ with the \textsc{small}, \textsc{base}, and \textsc{large} configurations (illustrated as S/B/L). All multiplexed models lie either on or very close to the Pareto frontier (shown in grey).
}
\vspace{-10pt}
\label{fig:pareto_plots}
\end{figure}

Pre-trained models typically have a performance-computational efficiency trade-off, with larger models having better performance but worse computational efficiency.
\muxplms{} offers a similar trade-off, with large $N$ leading to better throughput but lower performance.
To understand this trade-off, we plot the performance and throughput of \bert{} and \muxbert{} for different model sizes and draw the pareto-optimal envelope (Figure~\ref{fig:pareto_plots}).
For any model on the envelope, no model has both better accuracy and throughput. Users would only choose models on the envelope because for every model within the envelope, there always exists a model on the envelope which has \textit{both} better performance \textit{and} throughput.
We note that \textbf{all multiplexed models lie either on or very close to the Pareto frontier}, for both \token{} and \glue{} tasks. This suggests that given an accuracy threshold, \muxplm{} models will usually be faster than \standard{}. For instance, if we wanted the highest throughput model with a performance $\geq77\%$ on \glue{}, the optimal \bert{} model is the \textsc{small} configuration with a throughput of 2815 (in/s), but for the \muxbert{} model would be the $N=2$ with the \textsc{small} configuration, achieving a significantly higher throughput of 5539 (in/s). 

\subsection{Ensembling \muxplms{}}

\begin{table}[t]
\centering
\resizebox{\columnwidth}{!}{%
\begin{tabular}{lccccccc}
\toprule
\multirow{2}{*}{\textbf{Model}} & \multirow{2}{*}{\textbf{Mux ($N$)}} & \multicolumn{3}{c}{\textbf{MNLI}} & \multicolumn{3}{c}{\textbf{QQP}} \\ \cmidrule(lr){3-5} \cmidrule(lr){6-8}
                         &       & No Ens         & Ens  &  $\mathbf{\Delta}$         & No Ens        & Ens  & $\mathbf{\Delta}$        \\ \midrule
\multirow{3}{*}{\muxbert{}} &  \textbf{2}                   & 80.6        & \textbf{81.2} &   + 0.6    & 90.4       & \textbf{90.8} &   + 0.4   \\
 & \textbf{5}                           & 77.2        & \textbf{78.8} &  + 1.6    & 88.8      & \textbf{89.7}   &  + 0.9  \\
 & \textbf{10}                          & 73.6        & \textbf{74.8} &   + 1.2   & 86.9       & \textbf{87.7} &   + 0.8   \\ \midrule
 
\multirow{3}{*}{\muxelectratab{}} &  \textbf{2}                &  80.3         & \textbf{80.8} &   + 0.5    & 90.6       & \textbf{90.9} &   + 0.3   \\
 & \textbf{5}                           & 77.0       & \textbf{78.4} &  + 1.4    & 89.1       & \textbf{89.9}   &  + 0.8  \\
 & \textbf{10}                          & 74.6        & \textbf{76.0} &   + 1.4   & 87.6       & \textbf{88.3} &  + 0.7   \\ \bottomrule

\end{tabular}

}
\caption{Ensembling results for \muxbert{} and \muxelectra{} models for $N\in\{2,5,10\}$. \textit{Ens} denotes Ensembling. Ensembling improves performance for all the models, with the gains increasing with increasing N. This suggests that the multiplexing approach can be naturally adapted to load-balancing applications, where the ensembling strategy can be changed based on demand.}
\label{tab:ensemble}
\end{table}

As opposed to feeding $N$ different instances to \muxplms{} to improve throughput, we consider an alternate setting where we feed the same instance $N$ times and build an ensemble by averaging the $N$ class logits to make a single prediction. We randomly permute the batch, after duplicating the instance $N$ times, to prevent distribution shift. We use the \textsc{base} size models for $N\in\{2,5,10\}$ for both \muxbert{} and \muxelectra{} (Table~\ref{tab:ensemble}).
\textbf{The ensemble model does significantly better than the non-ensemble variant} on both MNLI and QQP for all values of $N$ (e.g., $1.6$ and $0.9$ points on $N=5$ \muxbert{} for the two tasks).
We note that the improvement over the non-ensemble variant ($\Delta$) is better for higher $N$, due to the larger ensemble size. This result shows that non-ensemble variants are faster but perform slightly worse, while the ensemble variant performs better but is slower. A spectrum of models lie between these two extremes, where only a fraction of the $N$ multiplexed representations can be ensembled, allowing users to trade off performance and speed.

\section{Analysis}
\label{sec:analysis}
\begin{figure*}[t]
    \centering
    \includegraphics[width=\textwidth]{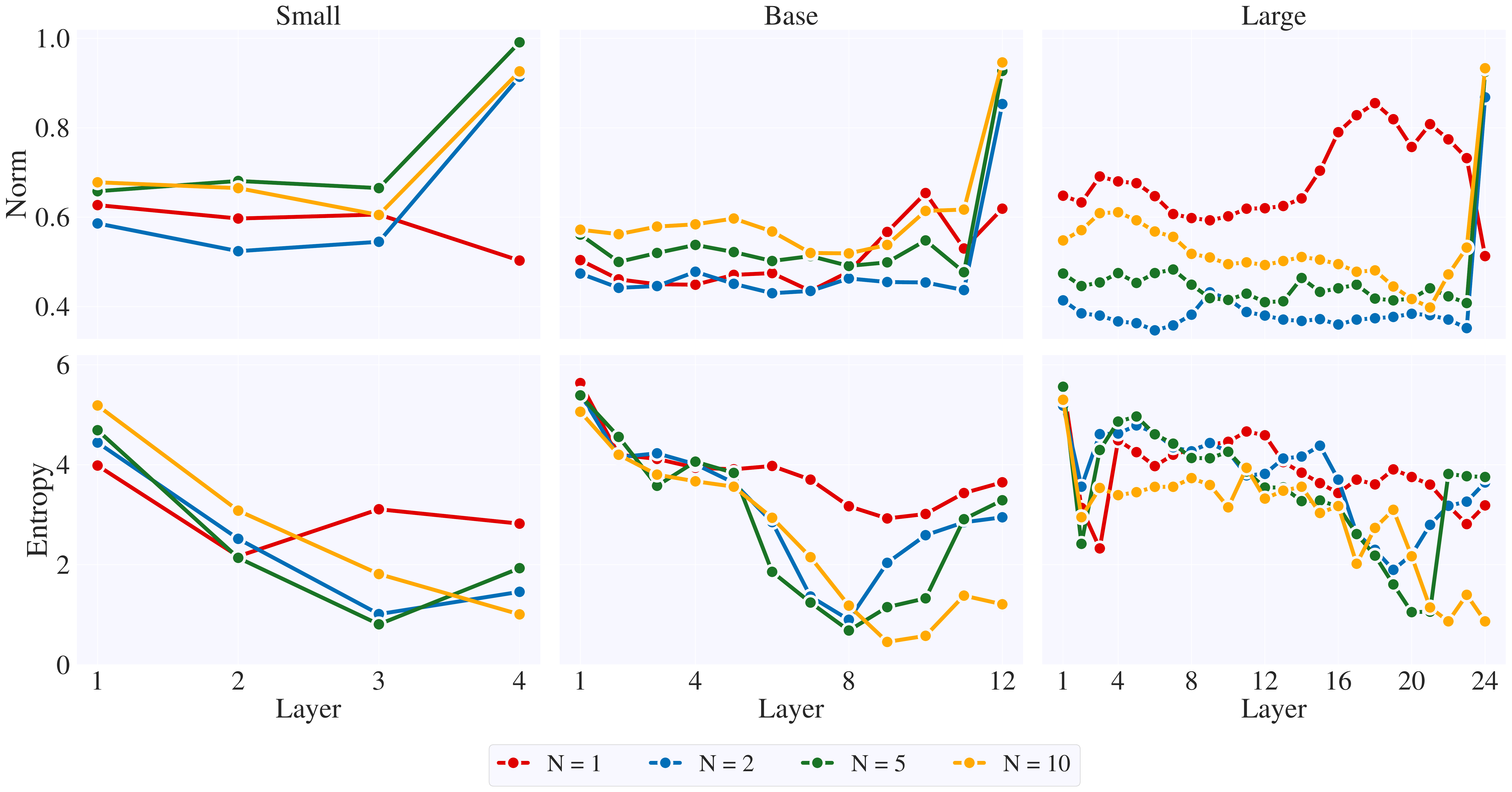}
    \caption{Comparing (Top) Layer-wise activation and (Bottom) attention entropy of \muxbert{} and \bert{}, for $N\in\{2,5,10\}$ across different configurations.  Activation norms tend to spike for \muxbert{} in the last layer and entropy of \muxbert{} is lower than \bert{} for higher layers.}
    \label{fig:activiation_norms}
\end{figure*}

\subsection{Ablation study}
\begin{table}[t]
    \centering
    \resizebox{\columnwidth}{!}{%
    \begin{tabular}{@{}clcccc@{}}
    \toprule
    \textbf{Mux (N)} & \textbf{Model} & \textbf{Mux} & \textbf{Demux} & \multicolumn{1}{c}{\textbf{GLUE}} & \multicolumn{1}{c}{\textbf{Token}} \\
    \midrule
    \multirow{3}{*}{\textbf{2}}  & \muxbert{} & \mainmux{} & \newdemux{} & 82.5\makeinvisible{/83.4} & 95.2\makeinvisible{/95.4} \\
                        & Ablation 1 & \mainmux{} & Prefix & \textbf{83.2\makeinvisible{/83.7}} & \textbf{95.3\makeinvisible{/95.4}} \\
                        & Ablation 2 & \newmux{} & \newdemux{} & 82.3\makeinvisible{/82.7}   & 95.3\makeinvisible{/95.3} \\
    \midrule
    \multirow{3}{*}{\textbf{5}}  & \muxbert{} & \mainmux{} & \newdemux{} & \textbf{80.3\makeinvisible{/80.9}} & 93.6\makeinvisible{/93.6} \\
                        & Ablation 1 & \mainmux{} & Prefix & 78.6\makeinvisible{/79.4} & \underline{38.9\makeinvisible{/39.5}}\\
                        & Ablation 2 & \newmux{} & \newdemux{} & 76.8\makeinvisible{/77.6}  &  \textbf{94.2\makeinvisible{/94.4}} \\
    
    \midrule
    \multirow{3}{*}{\textbf{10}} & \muxbert{} & \mainmux{} & \newdemux{}& \textbf{77.8\makeinvisible{/78.8}}  & 91.6\makeinvisible{/91.8}   \\
                        & Ablation 1 & \mainmux{} & Prefix & 76.6\makeinvisible{/77.2} &  \underline{25.6\makeinvisible{/26.5}}   \\
                        & Ablation 2 & \newmux{} & \newdemux{} & 76.0\makeinvisible{/77.0}  & \textbf{93.3\makeinvisible{/93.5}}  \\
    \bottomrule
    \end{tabular}%
    }
    \caption{Ablation analysis for \muxbert{} (base configuration) for $N\in\{2,5,10\}$. Across most configurations, the prefix demultiplexing variant performs worse than our proposed approach and fails to converge for token-level tasks for $N\in \{5,10\}$ (underlined numbers). The new contextual multiplexing variant (\newmux{}) outperforms \mainmux{} on token-level tasks.}
    \vspace{-10pt}
    \label{tab:ablation}
    \end{table}

We analyze multiplexing and demultiplexing components of \muxplms{} and report the results in Table~\ref{tab:ablation}. We consider two variants, one which uses the prefix demultiplexing proposed in~\tmux{} instead of our proposed \newdemux{} and another which uses \newmux{} multiplexing instead of \mainmux{}.
We note that Variant $1$, which uses prefix demultiplexing, performs worse than our \muxbert{}, other than for $N=2$.
In fact, Variant 1 does not converge for \token{} tasks for $N=5$ and $N=10$ and performs $1.7$ and $1.2$ points worse on \glue{} when compared to \muxbert{}.

Variant 2 uses \newmux{} multiplexing which takes into account other tokens present in the instance and also tokens present in the same position of other instances.
This variant performs better than \mainmux{} for \token{} tasks (almost over $1.7$ points on \token{}  for $N=10$) but performs worse for \glue{} tasks.
We believe that \newmux{} multiplexing's better performance in \token{} is because the model needs to make a prediction for every single position in the instance, which requires it to efficiently multiplex all token positions in the output.
However, for \glue{} tasks, the model needs to make a prediction only for the \texttt{[CLS]} token, for which \mainmux{} multiplexing suffices.

\subsection{Muxology: Analyzing hidden representations of multiplexed models}
\label{sec:muxology}
To understand the nature of representations being learned by \muxbert{} models, we analyze the absolute value of activations and entropy of the attention distribution across all the layers of the Transformer encoder, averaged over the evaluation split of WikiText-103~\cite{wikitext} (Figure~\ref{fig:activiation_norms}).
We report this analysis for different values of $N$ and for different model sizes.

\paragraph{1. Activation norms spike for \muxbert{} in the last layer.} Figure~\ref{fig:activiation_norms} (top) shows that activation norms spike in the last layer for multiplexed models.
We believe this is because the model is preparing for demultiplexing and is packing information from all $N$ instances, which makes the activations denser. We believe \muxbert{} has learned to efficiently encode multiple instances until the last layer where it needs to make independent predictions for them.

\paragraph{2. Entropy of the attention weights of \muxbert{} is lower than \bert{} for higher layers.}
Figure~\ref{fig:activiation_norms} (bottom) suggests that \muxbert{} tends to have lower entropy attention distributions on average as opposed to \bert{} for higher layers. This could be related to~\citet{deshpande2020guiding}'s observation of pre-trained models having peaky attention distributions in the higher layers, with small irregularities. Since the model implicitly has to use the same attention distribution for all the multiplexed instances, the peaky distribution gets reinforced and is further corroborated by higher $N$ having lower entropy in the final layer. We, therefore, believe that \muxbert{} has learned to create shared representations for multiple instances to effectively use the instance-independent attention distribution.

\subsection{Effect of data sampling strategies during inference}
\begin{table}[t]
\resizebox{\columnwidth}{!}{%
\begin{tabular}{@{}ccccccc@{}}
\toprule
\multirow{2}{*}{$\mathbf{N}$} & \multicolumn{3}{c}{\textbf{\muxelectra{}}} & \multicolumn{3}{c}{\textbf{\muxbert{}}} \\ \cmidrule(l){2-4} \cmidrule(l){5-7}
    & \textbf{Best ticket} & \textbf{Worst ticket} & $\Delta$ & \textbf{Best ticket} & \textbf{Worst ticket} & $\Delta$ \\ \midrule
\textbf{2} & 83.1 &  82.0 & 1.1 & 83.4 & 81.8 & 1.6  \\
\textbf{5} & 80.5 & 78.9 & 1.6 &  80.9 & 79.7 & 1.2 \\
\textbf{10} & 79.0 & 77.3 & 1.7 & 78.8  & 77.0 & 1.8\\ \bottomrule
\end{tabular}%
}
\caption{We consider 5 random seeds for every model variant, which can be viewed as lottery tickets as the seeds control the composition of N instances. We present the difference between the worst and the best-performing ticket across GLUE tasks and regularly see a $\geq1$ point difference.}
\label{tab:data_sampling}
\vspace{-20pt}
\end{table}

During inference, our \muxplms{} sample $N$ instances uniformly at random from the evaluation set.  
However, other data-sampling strategies such as clustering similar instances based on word-overlap could improve performance.
We explore the effect of composition of $N$ instances on the performance of \muxplms{} in Table~\ref{tab:data_sampling}.
For each model variant, we consider $5$ random seeds which can be viewed as lottery tickets~\cite{frankle2018lottery}.
Since the random seed controls the composition of $N$ instances, we measure the difference ($\Delta$) between the best-performing ticket and the worst-performing ticket and average the performance for all the \glue{} tasks.
$\Delta$ is consistently greater than $1$ point for all values of $N$ for both \muxelectra{} and \muxbert{}, and illustrates the importance of the composition of $N$ instances. An improved data sampling strategy could lead to improvements and we leave this to future work.

\section{Conclusion}
\label{sec:conclusion}

We introduce \muxplms{}, a class of high-throughput pre-trained language models trained with data multiplexing, a multi-input multi-output (MIMO) architecture.
Our \muxplms{} models, trained with novel MIMO modules, are competitive with state-of-the-art PLMs on several downstream tasks while achieving a many-fold increase in inference throughput.
\muxplms{}, similar to standard PLMs, can be fine-tuned on any downstream task to yield high-throughput, high-performance models.
We hope our work inspires future research in MIMO architectures for PLMs as a complementary efficiency paradigm to existing approaches.

\section*{Acknowledgements}
We gratefully acknowledge support from Google AI Princeton, where Vishvak Murahari was a student researcher for part of this work, and the Samsung GRO program. 
We thank Mengzhou Xia, Jens Tuyls, and Tianyu Gao, with special thanks to our espresso machine. 

\bibliography{99_references}
\bibliographystyle{acl_natbib}

\newpage
\appendix
\onecolumn
\section{Appendices}
\label{appendix:appendix}
\section{Pre-training Details}
\label{appendix:pre}
\addtolength{\tabcolsep}{0pt}
\begin{table*}[]
\begin{center}
\begin{tabular}{c c c c  c}
\toprule

\textbf{Hyperparameter} & \multicolumn{3}{c}{\textbf{\muxbert{}}} & \multicolumn{1}{c}{\textbf{\muxelectra{}}} \\
 \cmidrule(lr){2-4} \cmidrule(lr){5-5}
 & \textsc{Small} & \textsc{Base} & \textsc{Large} & \textsc{Base}  \\
\midrule
Number of layers & 4 & 12 & 24 & 12 \\
Hidden Size & 512 & 768 & 1024 & 768 \\
FFN intermediate hidden size & 2048 & 3072 & 4096 & 3072 \\
Attention heads & 8 & 12 & 16 & 12\\
Attention head size & 64 & 64 & 64 & 64 \\
Mask percent & 15 & 15 & 15 & N/A \\
Learning Rate Decay & Linear & Linear & Linear & Linear \\
Warmup steps & 10000 & 10000 & 10000 & 10000\\
Learning Rate & [1e-4, 5e-5] & [1e-4, 5e-5] & [1e-4, 5e-5] & [1e-4, 5e-5] \\
Adam $\epsilon$ & 1e-6 & 1e-6 & 1e-6 & 1e-6 \\
Adam $\beta_1$ & 0.9 & 0.9 & 0.9 & 0.9 \\
Adam $\beta_2$ & 0.999 & 0.999 & 0.999 & 0.999 \\
Attention Dropout & 0.1 & 0.1 & 0.1 & 0.1 \\
Dropout & 0.1 & 0.1 & 0.1 & 0.1\\
Batch Size & 256 & 256 & 256 & 256 \\
Sequence Length & 512 & 512 & 512  & 512\\
Train Steps & 1M & 1M & 1M  & 1M\\
\bottomrule
\end{tabular} 
\end{center}
\caption{Pre-train hyper-parameters for \muxbert{} and \muxelectra{} models. We only report results for the Base configuration for \muxelectra{} models.}
\label{tab:hyperpre}
\end{table*}

We report all pre-training related hyper-parameters in Table~\ref{tab:hyperpre}. We primarily use the HuggingFace Transformers implementations for BERT and ELECTRA based models. All pre-training experiments were run on 8 A100 GPUs with distributed training. We run a small hyper-parameter search over over two learning rates. All pre-trained models are primed with the token retrieval task introduced in~\citet{datamuxmurahari}. We train on the Wikipedia and Bookscorpus datasets for up to $10000$ training steps with a learning rate of $1e-4$, and with a sequence length of $512$. 

For \muxelectra{} models, we don't train a generator as in the original ELECTRA work, but only use uniform-random token replacement. This is similar to what was used in ablations in ELECTRA~\cite{clark2020electra}. The generator randomly replaces $15\%$ of tokens in the input with other tokens in the vocabulary.

\section{Fine-tuning Details}
\label{appendix:fine}

We report all the fine-tuning related hyper-parameters in Table~\ref{tab:hyperfine}. We run a small hyper-parameter search on the learning rate, batch size and number of training steps for different tasks. All models were trained with half-precision. We report numbers on the validation split. For GLUE tasks, we use the default metrics in~\citet{wang2018glue} and use F1 for the token-level tasks. All fine-tuning experiments were trained on 1 V100 GPU.
\paragraph{Speedup calculation} For all models, we calculate throughput (samples/second) on a single V100 GPU and report throughput gains with respect to the BERT$_{\textsc{base}}$ model. We calculate throughput by averaging across $3$ different trials (1 trial = 200 mini-batches) and use a batch size of $128$ and a sequence length of $128$ following prior work~\cite{cofi}. We measure throughput for sequence-classification tasks on QQP and measure throughput for token-level classification tasks on named entity recognition.

\begin{table*}[]
\begin{center}
\begin{tabular}{l l}
\toprule
\textbf{Hyperparameter} & \textbf{Value} \\
\midrule
Learning Rate & [2e-5, 5e-5] \\
Adam $\epsilon$ & 1e-8  \\
Adam $\beta_1$ & 0.9  \\
Adam $\beta_2$ & 0.999 \\
Learning rate decay & Linear \\
Warmup fraction & 0.1 \\
Attention Dropout & 0.1  \\
Dropout & 0.1  \\
Weight Decay & 0 \\
Batch Size & [32, 128] for \textsc{Small}/ \textsc{Base}, [16, 64] for \textsc{Large} \\
Train Steps & 2000 for RTE and WNLI \\ & 10000 for MRPC, COLA and STSB \\ & 20000 for NER, SST2, QNLI and POS \\ &  [20000, 100000] for MNLI and QQP \\
Sequence Length  & 128 \\
\bottomrule
\end{tabular} 
\end{center}
\caption{Fine-tune hyperparameters}
\label{tab:hyperfine}
\end{table*}

\section{Analysis details}
\label{appendix:analysis}

\subsection{Ensembling results setup}
We find that multiplexing the same instance by duplicating the instance N times leads to worse performance. This is likely because this input configuration is very out of distribution from what the multiplexed models are trained on. To address this, we randomly permute the instances in the batch after duplicating the instances N times. This ensures that the input to the multiplexer lies in a similar distribution to what the model was trained on.
\subsection{Muxology setup}
To analyze the hidden states of pre-trained \muxbert{} models at different layers, we take the average absolute value of hidden states and every layer for both multiplexed and baseline models, across different configurations. To analyze the entropies of the attention distributions at different layers, we calculate the attention distribution across different attention heads for each position in the sequence length. To measure how peaky the attention distribution is likely to be, we calculate the entropies of the attention distributions at all positions and average across all the positions and across all the attention heads to get the average entropy for all layers. We conduct this analysis on WikiText-103 and average across all the samples in the evaluation split.

\section{Task performance breakdown for all variants}
\label{appendix:taskperf}
\begin{table*}
\centering\resizebox{\textwidth}{!}{
\begin{tabular}{lccccccccccccccccccccccccccccccccccccccccccccccc}
\toprule
{Model Size} & {N} & {MNLI} & {QQP} & {QNLI} & {MRPC} & {WNLI} & {STSB} & {RTE} & {SST2} & {COLA} & {GLUE} & {GLUE$_{\small -\text{WNLI, COLA}}$} \\
\midrule
\multirow[c]{4}{*}{\textsc{Small}} & 1 & \textbf{77.86}$_{\pm 0.0}$ & \textbf{88.99}$_{\pm 0.0}$ & 84.00$_{\pm 0.0}$ & 77.70$_{\pm 0.0}$ & \textbf{56.34}$_{\pm 0.0}$ & \textbf{84.25}$_{\pm 0.0}$ & \textbf{62.45}$_{\pm 0.0}$ & \textbf{88.88}$_{\pm 0.0}$ & \textbf{43.48}$_{\pm 0.0}$ & \textbf{73.77} & \textbf{80.59} \\
 & 2 & 75.09$_{\pm 0.1}$ & 88.88$_{\pm 0.1}$ & \textbf{84.31}$_{\pm 0.2}$ & \textbf{79.75}$_{\pm 0.7}$ & 50.99$_{\pm 8.1}$ & 82.65$_{\pm 0.3}$ & 55.52$_{\pm 1.5}$ & 87.04$_{\pm 0.7}$ & 30.64$_{\pm 1.7}$ & {70.54} & {79.03} \\
 & 5 & 70.50$_{\pm 0.1}$ & 86.39$_{\pm 0.1}$ & 81.23$_{\pm 0.2}$ & 74.26$_{\pm 1.0}$ & \textbf{54.65}$_{\pm 3.3}$ & 79.90$_{\pm 0.2}$ & 58.56$_{\pm 1.9}$ & 82.57$_{\pm 0.3}$ & 12.78$_{\pm 1.6}$ & {66.76} & {76.20} \\
 & 10 & 61.98$_{\pm 0.1}$ & 80.85$_{\pm 0.1}$ & 63.47$_{\pm 0.3}$ & 70.69$_{\pm 0.9}$ & \textbf{56.62}$_{\pm 4.3}$ & 36.93$_{\pm 1.0}$ & 53.57$_{\pm 1.8}$ & 80.39$_{\pm 0.4}$ & 1.10$_{\pm 2.2}$ & {56.18} & {63.98} \\\midrule
\multirow[c]{4}{*}{\textsc{Base}} & 1 & \textbf{84.24}$_{\pm 0.0}$ & \textbf{91.19}$_{\pm 0.0}$ & \textbf{90.54}$_{\pm 0.0}$ & \textbf{87.75}$_{\pm 0.0}$ & \textbf{56.34}$_{\pm 0.0}$ & \textbf{89.18}$_{\pm 0.0}$ & \textbf{63.18}$_{\pm 0.0}$ & \textbf{91.74}$_{\pm 0.0}$ & \textbf{58.79}$_{\pm 0.0}$ & \textbf{79.22} & \textbf{85.40} \\
 & 2 & 80.59$_{\pm 0.1}$ & 90.36$_{\pm 0.1}$ & 88.17$_{\pm 0.1}$ & 83.77$_{\pm 1.4}$ & 50.70$_{\pm 7.0}$ & 85.84$_{\pm 0.1}$ & 58.19$_{\pm 1.6}$ & 90.62$_{\pm 0.6}$ & 55.61$_{\pm 1.6}$ & {75.98} & {82.51} \\
 & 5 & 77.18$_{\pm 0.2}$ & 88.79$_{\pm 0.1}$ & 85.58$_{\pm 0.1}$ & 80.10$_{\pm 0.6}$ & 53.52$_{\pm 2.5}$ & 84.28$_{\pm 0.2}$ & 59.13$_{\pm 1.2}$ & 86.88$_{\pm 0.4}$ & 12.33$_{\pm 2.4}$ & {69.75} & {80.28} \\
 & 10 & 73.62$_{\pm 0.3}$ & 86.94$_{\pm 0.1}$ & 82.08$_{\pm 0.3}$ & 78.63$_{\pm 0.6}$ & 52.68$_{\pm 6.0}$ & 81.62$_{\pm 0.2}$ & 58.27$_{\pm 2.4}$ & 83.44$_{\pm 0.6}$ & 0.00$_{\pm 0.0}$ & {66.36} & {77.80} \\\midrule
\multirow[c]{4}{*}{\textsc{Large}} & 1 & \textbf{85.79}$_{\pm 0.0}$ & \textbf{91.46}$_{\pm 0.0}$ & \textbf{92.29}$_{\pm 0.0}$ & 83.82$_{\pm 0.0}$ & \textbf{56.34}$_{\pm 0.0}$ & \textbf{89.53}$_{\pm 0.0}$ & \textbf{66.06}$_{\pm 0.0}$ & \textbf{91.40}$_{\pm 0.0}$ & \textbf{57.79}$_{\pm 0.0}$ & \textbf{79.39} & \textbf{85.76} \\
 & 2 & 83.23$_{\pm 0.2}$ & 90.85$_{\pm 0.1}$ & 90.66$_{\pm 0.2}$ & \textbf{84.90}$_{\pm 0.8}$ & \textbf{56.34}$_{\pm 0.0}$ & 88.22$_{\pm 0.2}$ & 59.21$_{\pm 0.9}$ & 91.38$_{\pm 0.4}$ & \textbf{57.89}$_{\pm 1.5}$ & {78.08} & {84.06} \\
 & 5 & 79.55$_{\pm 0.2}$ & 89.37$_{\pm 0.1}$ & 87.41$_{\pm 0.2}$ & 83.77$_{\pm 1.1}$ & \textbf{54.93}$_{\pm 0.0}$ & 85.86$_{\pm 0.3}$ & 57.26$_{\pm 2.0}$ & 88.65$_{\pm 0.7}$ & 46.66$_{\pm 0.9}$ & {74.83} & {81.70} \\
 & 10 & 35.45$_{\pm 0.0}$ & 63.18$_{\pm 0.0}$ & 50.54$_{\pm 0.0}$ & 68.38$_{\pm 0.0}$ & \textbf{56.90}$_{\pm 5.2}$ & 82.81$_{\pm 0.2}$ & 52.13$_{\pm 1.9}$ & 50.92$_{\pm 0.0}$ & 1.87$_{\pm 4.6}$ & {51.35} & {57.63} \\
\bottomrule
\end{tabular}}
\caption{We show the full GLUE results for \muxbert{}. We report the mean accuracy and standard deviation over 5 seeds. Extrema and values within their standard deviation are emphasized for each model size.}
\label{tab:full_glue_results_bert}
\end{table*}

\begin{table}
\centering\resizebox{\textwidth}{!}{
\begin{tabular}{lccccccccccccccccccccccccccccccccccccccccccccccc}
\toprule
{Model Size} & {N} & {MNLI} & {QQP} & {QNLI} & {MRPC} & {WNLI} & {STSB} & {RTE} & {SST2} & {COLA} & {GLUE} & {GLUE$_{-\text{WNLI, COLA}}$} \\
\midrule
\multirow[c]{4}{*}{\textsc{Small}} & 1 & \textbf{77.86} & \textbf{88.99} & {84.00} & {77.70} & {56.34} & \textbf{84.25} & \textbf{62.45} & \textbf{88.88} & \textbf{43.48} & \textbf{73.77} & \textbf{80.59} \\
 & 2 & {75.21} & \textbf{89.01} & \textbf{84.61} & \textbf{80.64} & {61.97} & {82.97} & {58.12} & {87.84} & {33.08} & {72.61} & {79.77} \\
 & 5 & {70.66} & {86.46} & {81.60} & {75.74} & {61.97} & {80.24} & {60.65} & {83.49} & {15.57} & {68.49} & {76.98} \\
 & 10 & {62.17} & {80.93} & {63.85} & {71.81} & \textbf{63.38} & {38.20} & {55.96} & {80.96} & {2.63} & {57.77} & {64.84} \\\midrule
\multirow[c]{4}{*}{\textsc{Base}} & 1 & \textbf{84.24} & \textbf{91.19} & \textbf{90.54} & \textbf{87.75} & {56.34} & \textbf{89.18} & \textbf{63.18} & \textbf{91.74} & \textbf{58.79} & \textbf{79.22} & \textbf{85.40} \\
 & 2 & {80.82} & {90.47} & {88.28} & {86.03} & \textbf{66.20} & {86.06} & {60.65} & {91.51} & {56.93} & {78.55} & {83.40} \\
 & 5 & {77.66} & {88.89} & {85.70} & {81.13} & {59.15} & {84.47} & {60.65} & {87.50} & {15.79} & {71.22} & {80.86} \\
 & 10 & {74.04} & {87.03} & {82.45} & {79.41} & {63.38} & {81.89} & {62.45} & {84.29} & {0.00} & {68.33} & {78.79} \\\midrule
\multirow[c]{4}{*}{\textsc{Large}} & 1 & \textbf{85.79} & \textbf{91.46} & \textbf{92.29} & {83.82} & {56.34} & \textbf{89.53} & \textbf{66.06} & {91.40} & {57.79} & \textbf{79.39} & \textbf{85.76} \\
 & 2 & {83.40} & {90.94} & {90.96} & \textbf{86.27} & {56.34} & {88.50} & {60.29} & \textbf{91.86} & \textbf{60.50} & {78.78} & {84.60} \\
 & 5 & {79.69} & {89.43} & {87.81} & {84.80} & {57.75} & {86.49} & {60.65} & {89.45} & {47.56} & {75.96} & {82.62} \\
 & 10 & {35.46} & {63.18} & {50.89} & {68.38} & \textbf{61.97} & {83.04} & {55.60} & {50.92} & {7.55} & {53.00} & {58.21} \\
\bottomrule
\end{tabular}}
\caption{We show the full GLUE results for \muxbert{}. We report the \textit{maximum} accuracy over 5 seeds. Extrema are emphasized.}
\label{tab:full_glue_max_results_bert}
\end{table}

\begin{table}
\centering\resizebox{\textwidth}{!}{
\begin{tabular}{lcccccccccccccccccccccccccccccccccccccccc}
\toprule
{N} & {MNLI} & {QQP} & {QNLI} & {MRPC} & {WNLI} & {STSB} & {RTE} & {SST2} & {COLA} & {GLUE} & {GLUE$_{-\text{WNLI, COLA}}$} \\
\midrule
1 & \textbf{81.49}$_{\pm 0.0}$ & \textbf{90.73}$_{\pm 0.0}$ & \textbf{89.73}$_{\pm 0.0}$ & 75.98$_{\pm 0.0}$ & \textbf{56.34}$_{\pm 0.0}$ & \textbf{87.73}$_{\pm 0.0}$ & 57.76$_{\pm 0.0}$ & \textbf{91.51}$_{\pm 0.0}$ & \textbf{56.79}$_{\pm 0.0}$ & \textbf{76.45} & {82.13} \\
2 & 80.29$_{\pm 0.2}$ & 90.58$_{\pm 0.1}$ & 88.39$_{\pm 0.2}$ & \textbf{83.73}$_{\pm 0.7}$ & \textbf{57.18}$_{\pm 2.1}$ & 86.80$_{\pm 0.1}$ & \textbf{58.77}$_{\pm 1.1}$ & 88.65$_{\pm 0.4}$ & 51.92$_{\pm 1.7}$ & {76.26} & \textbf{82.46} \\
5 & 76.99$_{\pm 0.2}$ & 89.08$_{\pm 0.0}$ & 85.40$_{\pm 0.3}$ & 80.25$_{\pm 1.6}$ & \textbf{56.90}$_{\pm 4.5}$ & 84.27$_{\pm 0.2}$ & 57.26$_{\pm 1.0}$ & 85.09$_{\pm 1.0}$ & 26.89$_{\pm 1.2}$ & {71.35} & {79.76} \\
10 & 74.62$_{\pm 0.2}$ & 87.63$_{\pm 0.1}$ & 82.70$_{\pm 0.2}$ & 77.89$_{\pm 0.7}$ & 50.99$_{\pm 4.9}$ & 81.96$_{\pm 0.5}$ & \textbf{59.86}$_{\pm 2.1}$ & 82.71$_{\pm 0.5}$ & 27.76$_{\pm 2.3}$ & {69.57} & {78.20} \\
\bottomrule
\end{tabular}}
\caption{We show the full GLUE results for \muxelectra{}$_{\textsc{BASE}}$. We report the mean accuracy and standard deviation over 5 seeds. Extrema and values within their standard deviation are emphasized for each model size.}
\label{tab:full_glue_results_electra}
\end{table}

\begin{table}
\centering\resizebox{\textwidth}{!}{
\begin{tabular}{lccccccccccccccccccccccccccccccccccccccccccccccc}
\toprule
{N} & {Retreival Rate} & {MNLI} & {QQP} & {QNLI} & {MRPC} & {WNLI} & {STSB} & {RTE} & {SST2} & {COLA} & {GLUE} & {GLUE$_{-\text{WNLI, COLA}}$} \\
\midrule
\multirow[c]{4}{*}{2} & 0.0 & 83.23$_{\pm 0.2}$ & 90.85$_{\pm 0.1}$ & \textbf{90.66}$_{\pm 0.2}$ & \textbf{84.90}$_{\pm 0.8}$ & \textbf{56.34}$_{\pm 0.0}$ & \textbf{88.22}$_{\pm 0.2}$ & \textbf{59.21}$_{\pm 0.9}$ & \textbf{91.38}$_{\pm 0.4}$ & \textbf{57.89}$_{\pm 1.5}$ & \textbf{78.08} & \textbf{84.06} \\
 & 0.1 & \textbf{83.55}$_{\pm 0.3}$ & \textbf{90.90}$_{\pm 0.1}$ & \textbf{90.58}$_{\pm 0.2}$ & \textbf{85.49}$_{\pm 1.1}$ & \textbf{56.34}$_{\pm 0.0}$ & \textbf{88.28}$_{\pm 0.2}$ & \textbf{57.76}$_{\pm 1.4}$ & 90.69$_{\pm 0.8}$ & \textbf{59.36}$_{\pm 1.4}$ & \textbf{78.11} & {83.89} \\
 & 0.2 & \textbf{83.50}$_{\pm 0.1}$ & \textbf{90.96}$_{\pm 0.1}$ & \textbf{90.69}$_{\pm 0.2}$ & \textbf{84.95}$_{\pm 0.5}$ & \textbf{56.34}$_{\pm 0.0}$ & \textbf{88.28}$_{\pm 0.2}$ & \textbf{58.34}$_{\pm 1.6}$ & 90.69$_{\pm 0.5}$ & \textbf{59.17}$_{\pm 1.5}$ & \textbf{78.10} & {83.92} \\
 & 0.5 & \textbf{83.41}$_{\pm 0.2}$ & \textbf{90.91}$_{\pm 0.0}$ & 90.47$_{\pm 0.1}$ & \textbf{85.25}$_{\pm 0.5}$ & \textbf{56.34}$_{\pm 0.0}$ & 88.02$_{\pm 0.1}$ & \textbf{59.35}$_{\pm 1.6}$ & 89.52$_{\pm 0.6}$ & \textbf{59.41}$_{\pm 2.0}$ & \textbf{78.08} & {83.85} \\\midrule
\multirow[c]{4}{*}{5} & 0.0 & \textbf{79.55}$_{\pm 0.2}$ & \textbf{89.37}$_{\pm 0.1}$ & \textbf{87.41}$_{\pm 0.2}$ & \textbf{83.77}$_{\pm 1.1}$ & 54.93$_{\pm 0.0}$ & \textbf{85.86}$_{\pm 0.3}$ & \textbf{57.26}$_{\pm 2.0}$ & \textbf{88.65}$_{\pm 0.7}$ & \textbf{46.66}$_{\pm 0.9}$ & \textbf{74.83} & \textbf{81.70} \\
 & 0.1 & \textbf{79.49}$_{\pm 0.1}$ & \textbf{89.34}$_{\pm 0.1}$ & \textbf{87.25}$_{\pm 0.3}$ & 81.81$_{\pm 1.3}$ & 53.24$_{\pm 1.6}$ & \textbf{85.80}$_{\pm 0.2}$ & \textbf{55.60}$_{\pm 2.4}$ & \textbf{88.19}$_{\pm 0.7}$ & \textbf{47.60}$_{\pm 1.0}$ & {74.26} & {81.07} \\
 & 0.2 & \textbf{79.37}$_{\pm 0.1}$ & \textbf{89.42}$_{\pm 0.1}$ & \textbf{87.23}$_{\pm 0.3}$ & 82.40$_{\pm 1.1}$ & 54.93$_{\pm 0.0}$ & \textbf{85.85}$_{\pm 0.2}$ & \textbf{55.38}$_{\pm 2.6}$ & 87.84$_{\pm 0.8}$ & 43.58$_{\pm 1.2}$ & {74.00} & {81.07} \\
 & 0.5 & 79.24$_{\pm 0.1}$ & 89.30$_{\pm 0.1}$ & 87.21$_{\pm 0.3}$ & 82.06$_{\pm 1.7}$ & \textbf{56.34}$_{\pm 0.0}$ & \textbf{85.97}$_{\pm 0.2}$ & 52.27$_{\pm 4.0}$ & \textbf{88.58}$_{\pm 0.6}$ & \textbf{47.01}$_{\pm 2.3}$ & {74.22} & {80.66} \\\midrule
\multirow[c]{4}{*}{10} & 0.0 & \textbf{35.45}$_{\pm 0.0}$ & \textbf{63.18}$_{\pm 0.0}$ & \textbf{50.54}$_{\pm 0.0}$ & \textbf{68.38}$_{\pm 0.0}$ & \textbf{56.90}$_{\pm 5.2}$ & \textbf{82.81}$_{\pm 0.2}$ & 52.13$_{\pm 1.9}$ & 50.92$_{\pm 0.0}$ & \textbf{1.87}$_{\pm 4.6}$ & \textbf{51.35} & \textbf{57.63} \\
 & 0.1 & \textbf{35.45}$_{\pm 0.0}$ & \textbf{63.18}$_{\pm 0.0}$ & \textbf{50.65}$_{\pm 0.2}$ & \textbf{68.38}$_{\pm 0.0}$ & \textbf{54.93}$_{\pm 5.0}$ & 4.45$_{\pm 1.5}$ & 51.48$_{\pm 2.4}$ & 50.92$_{\pm 0.0}$ & \textbf{1.34}$_{\pm 1.8}$ & {42.31} & {46.36} \\
 & 0.2 & \textbf{35.45}$_{\pm 0.0}$ & \textbf{63.18}$_{\pm 0.0}$ & 50.21$_{\pm 0.5}$ & \textbf{68.43}$_{\pm 0.8}$ & \textbf{54.65}$_{\pm 4.2}$ & 0.23$_{\pm 1.5}$ & 52.35$_{\pm 2.0}$ & \textbf{51.72}$_{\pm 0.4}$ & \textbf{0.29}$_{\pm 2.7}$ & {41.83} & {45.94} \\
 & 0.5 & \textbf{35.45}$_{\pm 0.0}$ & \textbf{63.18}$_{\pm 0.0}$ & 50.43$_{\pm 0.4}$ & \textbf{68.38}$_{\pm 0.0}$ & \textbf{56.06}$_{\pm 0.6}$ & 82.01$_{\pm 0.6}$ & \textbf{52.71}$_{\pm 0.0}$ & 50.92$_{\pm 0.0}$ & \textbf{1.51}$_{\pm 1.7}$ & {51.18} & \textbf{57.58} \\
\bottomrule
\end{tabular}}
\caption{GLUE results for \muxbert{}$_{\textsc{LARGE}}$ when using a retrieval auxiliary objective during MLM pretraining with different trade-off rates to the MLM objective. We report the average accuracy over 5 seeds. Extrema and values within their standard deviation are emphasized for each value of N.}
\label{tab:full_glue_results_bert_retreival_rate_ablations}
\end{table}

\begin{table}
\centering\resizebox{\textwidth}{!}{
\begin{tabular}{lccccccccccccccccccccccccccccccccccccccccccccccc}
\toprule
{N} & {Mux Strategy} & {MNLI} & {QQP} & {QNLI} & {MRPC} & {WNLI} & {STSB} & {RTE} & {SST2} & {COLA} & {GLUE} & {GLUE$_{-\text{WNLI, COLA}}$} \\
\midrule
\multirow[c]{3}{*}{2} & \muxbert{} & 80.59$_{\pm 0.1}$ & 90.36$_{\pm 0.1}$ & 88.17$_{\pm 0.1}$ & 83.77$_{\pm 1.4}$ & 50.70$_{\pm 7.0}$ & 85.84$_{\pm 0.1}$ & 58.19$_{\pm 1.6}$ & 90.62$_{\pm 0.6}$ & \textbf{55.61}$_{\pm 1.6}$ & {75.98} & {82.51} \\
 & DataMUX & \textbf{81.64}$_{\pm 0.2}$ & \textbf{90.67}$_{\pm 0.1}$ & 88.39$_{\pm 0.2}$ & \textbf{84.17}$_{\pm 0.4}$ & \textbf{56.34}$_{\pm 0.0}$ & \textbf{86.36}$_{\pm 0.2}$ & \textbf{60.87}$_{\pm 0.7}$ & 90.50$_{\pm 0.4}$ & 53.74$_{\pm 1.0}$ & \textbf{76.96} & \textbf{83.23} \\
 & Attention & 81.32$_{\pm 0.2}$ & \textbf{90.65}$_{\pm 0.0}$ & \textbf{88.77}$_{\pm 0.1}$ & 80.88$_{\pm 0.6}$ & \textbf{56.34}$_{\pm 0.0}$ & \textbf{86.25}$_{\pm 0.1}$ & 56.90$_{\pm 1.2}$ & \textbf{91.06}$_{\pm 0.2}$ & 47.15$_{\pm 1.1}$ & {75.48} & {82.26} \\\midrule
\multirow[c]{3}{*}{5} & \muxbert{} & \textbf{77.18}$_{\pm 0.2}$ & 88.79$_{\pm 0.1}$ & \textbf{85.58}$_{\pm 0.1}$ & \textbf{80.10}$_{\pm 0.6}$ & 53.52$_{\pm 2.5}$ & \textbf{84.28}$_{\pm 0.2}$ & \textbf{59.13}$_{\pm 1.2}$ & \textbf{86.88}$_{\pm 0.4}$ & 12.33$_{\pm 2.4}$ & {69.75} & \textbf{80.28} \\
 & DataMUX & 76.32$_{\pm 0.1}$ & \textbf{89.13}$_{\pm 0.1}$ & 84.22$_{\pm 0.3}$ & 78.38$_{\pm 0.9}$ & \textbf{59.44}$_{\pm 3.5}$ & 81.78$_{\pm 0.4}$ & 54.15$_{\pm 1.3}$ & 86.17$_{\pm 0.4}$ & 28.32$_{\pm 0.8}$ & \textbf{70.88} & {78.59} \\
 & Attention & \textbf{77.16}$_{\pm 0.1}$ & 88.71$_{\pm 0.0}$ & 84.33$_{\pm 0.1}$ & 70.49$_{\pm 0.6}$ & 54.08$_{\pm 3.2}$ & 80.37$_{\pm 0.3}$ & 54.44$_{\pm 2.5}$ & 81.95$_{\pm 0.3}$ & \textbf{34.67}$_{\pm 1.2}$ & {69.58} & {76.78} \\\midrule
\multirow[c]{3}{*}{10} & \muxbert{} & \textbf{73.62}$_{\pm 0.3}$ & 86.94$_{\pm 0.1}$ & 82.08$_{\pm 0.3}$ & \textbf{78.63}$_{\pm 0.6}$ & 52.68$_{\pm 6.0}$ & 81.62$_{\pm 0.2}$ & \textbf{58.27}$_{\pm 2.4}$ & \textbf{83.44}$_{\pm 0.6}$ & 0.00$_{\pm 0.0}$ & {66.36} & \textbf{77.80} \\
 & DataMUX & 72.74$_{\pm 0.1}$ & 87.88$_{\pm 0.1}$ & \textbf{82.28}$_{\pm 0.2}$ & 77.30$_{\pm 0.5}$ & \textbf{56.34}$_{\pm 0.0}$ & 78.07$_{\pm 0.4}$ & 55.31$_{\pm 1.2}$ & 82.36$_{\pm 0.3}$ & 13.56$_{\pm 3.0}$ & {67.32} & {76.56} \\
 & Attention & 71.83$_{\pm 0.2}$ & \textbf{88.00}$_{\pm 0.0}$ & 81.46$_{\pm 0.2}$ & 73.53$_{\pm 0.5}$ & 53.24$_{\pm 5.4}$ & \textbf{82.95}$_{\pm 0.2}$ & 52.71$_{\pm 0.0}$ & 81.28$_{\pm 0.4}$ & \textbf{32.84}$_{\pm 0.6}$ & \textbf{68.65} & {75.97} \\
\bottomrule
\end{tabular}}
\caption{GLUE results for \muxbert{}$_{\textsc{base}}$ using alternative multiplexing-demultiplexing strategies. We report the average accuracy over 5 seeds. Extrema and values within their standard deviation are emphasized for each value of N.}
\label{tab:full_glue_results_bert_multiplexing_ablations}
\end{table}

\begin{table}
\centering\resizebox{\textwidth}{!}{
\begin{tabular}{lcccccccccccccccccccccccccccccccccccccccccccccc}
\toprule
{Model Size} & {N} & {MNLI} & {QQP} & {QNLI} & {MRPC} & {WNLI} & {STSB} & {RTE} & {SST2} & {COLA} & {GLUE} & {GLUE$_{-\text{WNLI, COLA}}$} \\
\midrule
\multirow[c]{3}{*}{\textsc{Small}} & 2 & \textbf{61.48}$_{\pm 0.2}$ & \textbf{80.33}$_{\pm 0.0}$ & \textbf{60.05}$_{\pm 0.2}$ & \textbf{68.43}$_{\pm 0.5}$ & \textbf{56.34}$_{\pm 0.0}$ & \textbf{15.02}$_{\pm 0.4}$ & \textbf{51.12}$_{\pm 0.6}$ & \textbf{79.75}$_{\pm 0.3}$ & \textbf{8.22}$_{\pm 0.7}$ & \textbf{53.42} & \textbf{59.45} \\
 & 5 & 58.35$_{\pm 0.2}$ & 77.50$_{\pm 0.1}$ & 57.17$_{\pm 0.3}$ & \textbf{68.38}$_{\pm 0.0}$ & \textbf{56.34}$_{\pm 0.0}$ & 11.31$_{\pm 0.3}$ & \textbf{51.70}$_{\pm 1.3}$ & 77.78$_{\pm 0.3}$ & 6.02$_{\pm 0.7}$ & {51.62} & {57.46} \\
 & 10 & 53.63$_{\pm 0.2}$ & 77.03$_{\pm 0.1}$ & 51.22$_{\pm 0.3}$ & \textbf{68.38}$_{\pm 0.0}$ & \textbf{57.46}$_{\pm 6.3}$ & 12.40$_{\pm 1.3}$ & \textbf{52.35}$_{\pm 2.7}$ & 50.92$_{\pm 0.0}$ & 0.00$_{\pm 0.0}$ & {47.04} & {52.28} \\\midrule
\multirow[c]{3}{*}{\textsc{Base}} & 2 & \textbf{63.29}$_{\pm 0.3}$ & \textbf{81.42}$_{\pm 0.1}$ & \textbf{60.35}$_{\pm 0.4}$ & 68.38$_{\pm 0.2}$ & 56.90$_{\pm 5.8}$ & \textbf{17.65}$_{\pm 1.0}$ & 51.19$_{\pm 1.7}$ & \textbf{80.78}$_{\pm 0.5}$ & \textbf{9.62}$_{\pm 1.5}$ & \textbf{54.40} & \textbf{60.44} \\
 & 5 & 60.67$_{\pm 0.2}$ & 79.42$_{\pm 0.1}$ & 59.77$_{\pm 0.2}$ & \textbf{69.61}$_{\pm 0.8}$ & 53.80$_{\pm 7.3}$ & 14.92$_{\pm 1.8}$ & \textbf{52.71}$_{\pm 0.8}$ & \textbf{81.15}$_{\pm 0.6}$ & \textbf{10.35}$_{\pm 1.7}$ & {53.60} & {59.75} \\
 & 10 & 59.07$_{\pm 0.2}$ & 78.22$_{\pm 0.1}$ & 57.99$_{\pm 0.5}$ & 68.38$_{\pm 0.0}$ & \textbf{60.28}$_{\pm 3.0}$ & 11.83$_{\pm 0.6}$ & \textbf{53.07}$_{\pm 1.1}$ & 78.35$_{\pm 1.1}$ & 7.40$_{\pm 1.7}$ & {52.73} & {58.13} \\\midrule
\multirow[c]{3}{*}{\textsc{Large}} & 2 & \textbf{64.64}$_{\pm 0.2}$ & \textbf{82.10}$_{\pm 0.1}$ & \textbf{60.21}$_{\pm 0.2}$ & \textbf{69.95}$_{\pm 0.9}$ & \textbf{56.34}$_{\pm 0.0}$ & \textbf{21.62}$_{\pm 0.4}$ & 52.71$_{\pm 0.0}$ & \textbf{80.34}$_{\pm 0.9}$ & \textbf{8.72}$_{\pm 2.1}$ & \textbf{55.18} & \textbf{61.65} \\
 & 5 & 60.78$_{\pm 0.3}$ & 78.56$_{\pm 0.1}$ & \textbf{60.19}$_{\pm 0.3}$ & \textbf{69.51}$_{\pm 0.5}$ & \textbf{56.34}$_{\pm 0.0}$ & 17.33$_{\pm 1.1}$ & 52.71$_{\pm 0.0}$ & 78.28$_{\pm 0.8}$ & \textbf{10.63}$_{\pm 2.7}$ & {53.81} & {59.62} \\
 & 10 & 48.79$_{\pm 0.6}$ & 68.41$_{\pm 0.1}$ & 55.76$_{\pm 0.8}$ & 68.58$_{\pm 0.6}$ & \textbf{58.59}$_{\pm 3.3}$ & 8.38$_{\pm 1.1}$ & \textbf{54.95}$_{\pm 0.9}$ & 64.82$_{\pm 1.0}$ & 3.48$_{\pm 3.9}$ & {47.97} & {52.81} \\
\bottomrule
\end{tabular}}
\caption{GLUE results for \tmux{} with the original training recipe and implementation from ~\citet{datamuxmurahari}. We report the average accuracy and standard deviation over 5 seeds. Extrema and values within their standard deviation are emphasized for each model size.}
\label{tab:full_glue_results_datamux_orig}
\end{table}


\end{document}